\documentclass[10pt]{article} 
\usepackage[preprint]{tmlr}

\usepackage{amsmath,amsfonts,bm}

\def\eqref#1{equation~\ref{#1}}

\def\1{\bm{1}}

\DeclareMathAlphabet{\mathsfit}{\encodingdefault}{\sfdefault}{m}{sl}
\SetMathAlphabet{\mathsfit}{bold}{\encodingdefault}{\sfdefault}{bx}{n}

\usepackage{graphicx}
\usepackage{svg}
\usepackage{nicefrac}
\usepackage{wrapfig, booktabs}
\usepackage{algorithm,algpseudocode}
\usepackage{amsmath}
\usepackage{amsfonts}
\usepackage{amssymb}
\usepackage{amsthm}

\usepackage{bbm}
\usepackage{enumitem}
\usepackage{chngcntr}
\usepackage{comment}
\usepackage{color, colortbl}

\definecolor{LightCyan}{rgb}{0.9,0.96,1}
\definecolor{my_green}{RGB}{0,104,56}

\usepackage{filecontents}

\usepackage{hyperref}
\usepackage{url}

\title{The Impact of Semantic Pairs on Self-Supervised Representation Learning}

\author{\name Mohammad Alkhalefi \email m.alkhalefi1.21@abdn.ac.uk\\
      \addr Department of Computing Science\\
      University of Aberdeen
      \AND
      \name Georgios Leontidis \email georgios.leontidis@uit.no\\
      \addr Department of Physics and Technology\\
      UiT The Arctic University of Norway
      \AND
      \name Mingjun Zhong \email mingjun.zhong@abdn.ac.uk\\
      \addr Department of Computing Science \\
      University of Aberdeen\\
      }

\def\month{MM}  
\def\year{YYYY} 
\def\openreview{\url{https://openreview.net/forum?id=XXXX}} 

\begin{document}

\maketitle

\begin{abstract}
  Instance discrimination learns visual representations by treating different augmented views of the same image as positive pairs. While this encourages invariance to handcrafted transformations, same-image positives can preserve nuisance correlations such as background, texture, illumination, and object-specific details. Semantic positive pairs, i.e., different same-class instances, may reduce these correlations by presenting objects across diverse contexts. However, previous studies often combine semantic pairs with augmented positives or false neighbors (i.e., incorrectly mapped semantic pairs), making it difficult to isolate the effect of semantic pairing. We present a controlled empirical study of semantic positive pairs for self-supervised representation learning. From ImageNet-1K, we construct two matched subsets: an augmented-pair baseline and a manually curated semantic-pair dataset with the same class composition and training-pair count. We use these datasets to compare representative contrastive and non-contrastive SSL methods under matched training conditions. Across transfer learning and object detection evaluations, semantic-pair pretraining consistently improves generalisation over augmented-pair pretraining. Additional ablations show that semantic pairs induce invariances beyond the standard transformation pipeline. Among the evaluated methods, contrastive learning benefits most strongly from semantic pairs, with SimCLR showing the largest relative improvement. These results clarify the role of semantic positive pairs in SSL and provide guidance for selecting and designing frameworks that can exploit semantic pair information effectively.
   
\end{abstract}

\section{Introduction}
\section{Introduction}
Instance discrimination is a self-supervised learning (SSL) approach that allows the model to learn useful object representations without manual data annotation \citep{ZHANG2025131409}. These approaches treat each instance in the dataset as a class and discriminate it from the other instances. To achieve this, they attract the two augmented views (i.e., positive pairs) of the same instance in the embedding space and use different techniques to avoid collapse to the trivial representation  \citep{bardes2021vicreg,chen2020simple,chen2020improved,chen2021exploring}.  Although data transformations, such as geometric (e.g., random cropping, rotation) and appearance-based (e.g., colour jitter, blur), help the model learn invariant representations against nuisances, relying solely on transformations introduces limitations that may constrain generalisation to unseen data  \citep{alkhalefi2023semantic,tian2020makes,xiao2020should}.  Recent work on data curation for visual contrastive learning similarly argues that single-instance positive-pair construction can limit data diversity, since stochastic augmentations may fail to capture important variations such as viewpoint changes, object deformations, and semantically similar same-class instances \citep{desai2025survey}.
To better understand the limitations of data transformation, we must realise how instance discrimination approaches learn object representations from positive views.  In Figure 1(a), the learnt representations will capture the shared information between the two views ($x^1$ and $x^2$) \citep{alkhalefi2024leoclr,tian2020makes,zhang2022leverage,meehan2023ssl}. As shown in the Figure, the two views still have similar backgrounds and similar door stickers on the truck. This might lead to inappropriate relationships between the background and foreground, as well as other details, such as the ‘sticker’, which reduces the model's ability to generalise to the unseen data  \citep{meehan2023ssl,zeng2024contrastive}.  The solution to this problem is to reduce the shared information between the two views while keeping the relevant task information intact (i.e., truck semantic features) \citep{tian2020makes}. Semantic pairs (i.e., two different instances belonging to the same class) place the object in different contexts, allowing the model to focus on the relevant shared information and downweight the nuisance information.  In Figure 1(b), the aim is to learn a model that captures the general representation of the tow truck by exposing the model to two views containing ‘tow trucks’ but in different contexts. Since the captured representation relies on shared information across the two views (i.e., the same information consistently present in both views), placing the truck in different contexts encourages the model to focus on the relevant task information (i.e., learn the truck representation) and discard the irrelevant information needed to recognize the truck (i.e., background and sticker on the door). This leads the model to create a general representation of the tow truck based on the semantic features of the truck (i.e., cab, lights, tow part, etc.) while reducing the associative relationship between the ‘tow truck’ and other irrelevant information present in the image. In other words, the model creates a general representation of the truck; this representation captures the semantic features of the truck and is invariant to nuisance information, enabling it to recognise the truck in different contexts during downstream tasks. 

\begin{figure*}[h]
\begin{center}
\includegraphics[width=0.65\textwidth]{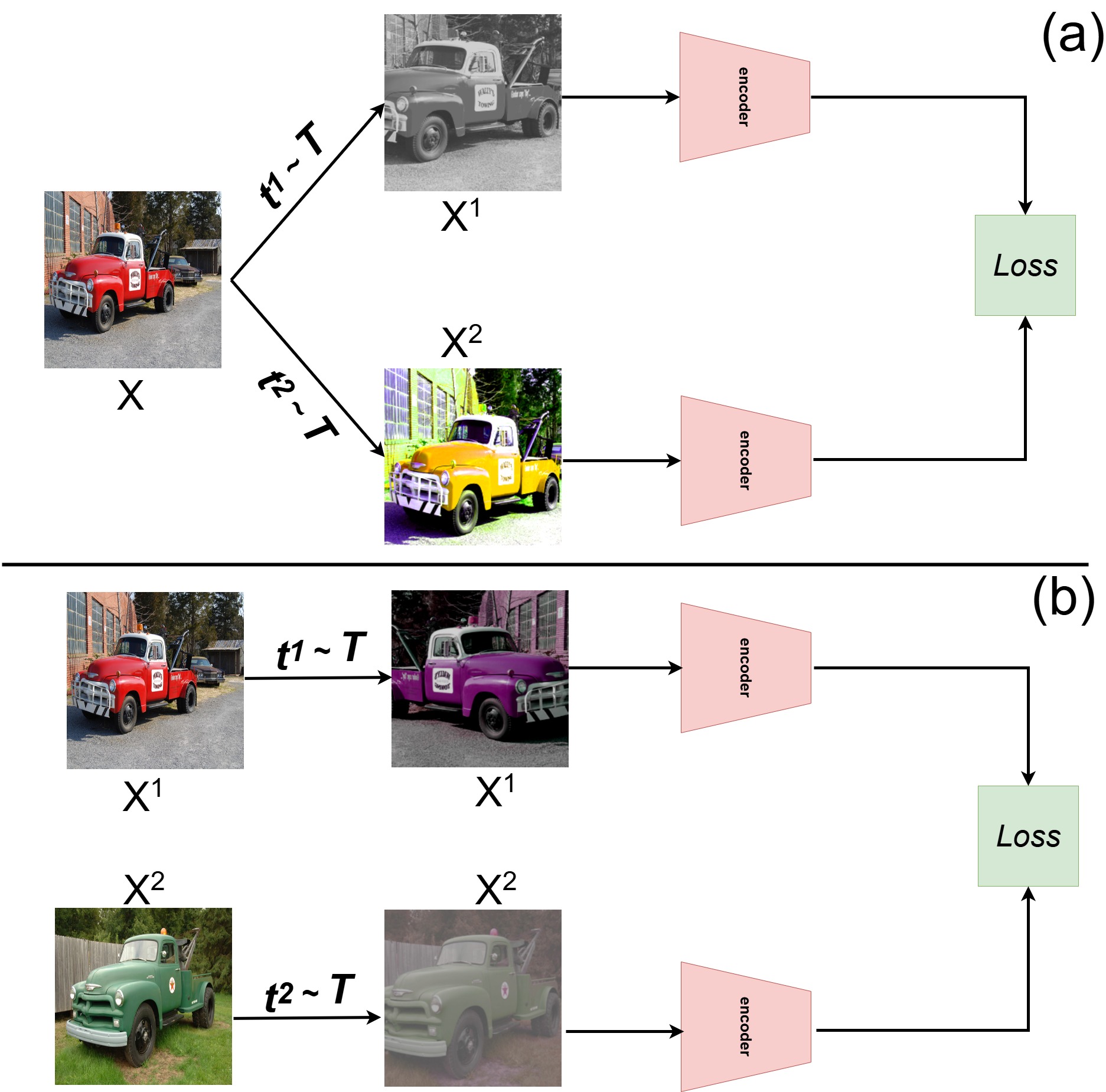}

\end{center}
   
   \caption{This diagram illustrates how instance discrimination approaches treat (a) augmented pairs (i.e., two views for the same instance) and (b) semantic pairs (i.e., two instances belonging to the same category) during representation learning.}
\label{fig:figure1}
\end{figure*}

Training the model on semantic pairs using self-supervised instance discrimination approaches exposes it to various contextual instantiations of an object, which enhances the learning of the object's high-level features (i.e., semantic features). Consequently, it improves model generalization and enhances the model's performance on unseen data. This study demonstrates how semantic pairs enhance the representation learning of instance discrimination approaches and conducts experiments to validate the theory behind using semantic pairs. Pretraining on semantic pairs guides the learning process of visual representations towards greater invariance to factors such as occlusion, background, pattern, and illumination, supporting the goal of learning general representations.

\textbf{Occlusion Invariance.} 
Occlusion invariance means the model can recognize objects even when parts of them are hidden or obscured. This is a crucial property for robust object recognition, as objects in real-world scenes are often partially occluded by other objects. The semantic pairs provide the model with views for the same category but with different types of occlusion.  This allows the model to create representations for the shared information of the object between the two views, treat the other irrelevant information as noise, and discard it \citep{purushwalkam2020demystifying,alkhalefi2024leoclr,zeng2024contrastive,wang2022contrastive}. For example, in Figure 2 (occlusion '2'), the first view shows a bird partially blocked by tree branches, while another view shows a complete bird. When the objective function forces the model to develop a similar representation for both views, it captures the remaining semantic features of the birds (i.e., head, body, and tail) present in both views and dismisses the tree branches that appear in only one view. Exposing the model to various instances of natural occlusions (different objects hiding different parts of the bird) can encourage learning a more general representation of "bird" that is not tied to specific viewpoints or occlusion patterns.

\textbf{Background invariance.}
Background invariance means the model can recognize objects in different contexts or backgrounds. In SSL instance discrimination, the model treats each instance as a class of its own and learns the representation of the object based on the shared information between the two augmented views. This might lead to memorizing the images and associating a relation between the background and foreground object despite no relation between them \citep{meehan2023ssl,zeng2024contrastive}.
In this case, the generalization of representations is diminished, which limits the model's ability to recognize objects with different backgrounds in downstream tasks. Semantic pairs represent the same object from two views but within different contexts. By varying the background between paired views, the instance discrimination model focuses on the semantic features of the object depicted in the two views, treating the background as less relevant for object recognition \citep{zeng2024contrastive}. This breaks irrelevant relations between foreground and background and allows the model to capture better semantic features for the object, which leads to recognizing the object in different contexts and improving the performance of the models on different downstream tasks. For example, in Figure 2 (Background '1'), the model is presented with two views of the birdhouse, each with a distinct background. This encourages the model to focus on the semantic features of the birdhouse and reduce the influence of the background, as it is less relevant for recognizing the birdhouse. Thus, the model learns to be invariant to variations in background (i.e., it can recognize the object in the downstream task, regardless of the background). 

\textbf{Abstract representation.}
In real-world scenarios, identical objects, such as a brand's logo and decorative paint designs, can exhibit varying surface patterns. Training a model on semantically paired data creates a generalized representation for the object by leveraging the common information found in these two distinct views. This allows the model to recognize the object in various downstream tasks, regardless of its pattern.  For example, in Figure 2 (pattern '2'), the model is given two aeroplane images, each featuring a different brand’s logo (i.e., the Airways brand). The model will capture the semantic features of the aeroplane (i.e., tail, wings, fuselage, etc.) that exist in both views to create a general representation of the aeroplane, treating the logos as noise and discarding them. This enables the model to recognise the aeroplane with a different pattern in downstream tasks, known as pattern invariance.

\textbf{Illumination invariance.}
Since lighting conditions change in the real world (e.g., dim indoor lights, outdoor lights, shadows, reflections), we want models to recognize objects regardless of illumination. This is called illumination invariance. To achieve that, we need to train the model on the same object with different illumination \citep{xiao2024generalization}. Semantic pairs enable the model to train on the same object under varying illumination conditions. For example, if a model is trained on two views of the same object e.g., train, each with different lighting, as shown in Figure 2 (illumination '3'), it is encouraged to focus on the train's semantic features rather than the illumination. Because the illumination varies between the two views, the model learns to be less sensitive to it, effectively treating the illumination differences as less important for object recognition. This helps create a representation of the train that is more invariant to variations in illumination. 

\begin{figure*}[h]
\begin{center}
\includegraphics[width=0.6\textwidth]{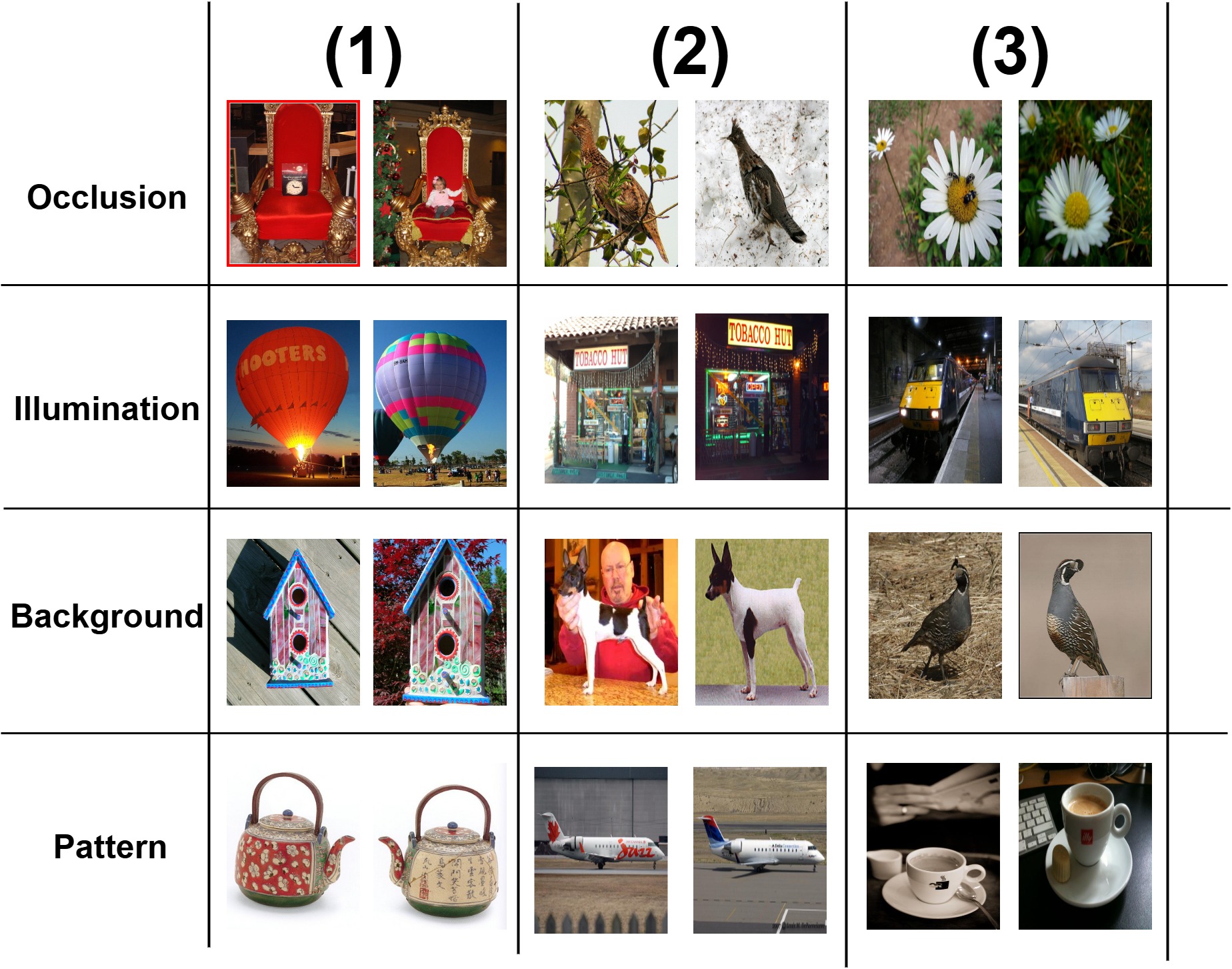}

\end{center}
   
   \caption{Examples from the semantic pairs dataset illustrate placing objects in diverse real-world context scenes.}
\label{fig:figure2}
\end{figure*}

Our contributions are as follows:


\begin{itemize}
    \item     We elucidate the mechanism by which semantic pairing induces representation invariance that leads to enhanced generalization capabilities.
    \item     We conduct a controlled empirical study using a custom-built dataset, allowing us to uniquely isolate the impact of semantic pairs from standard positive pairs and false neighbors during self-supervised learning.
    \item     We compare representative contrastive and non-contrastive SSL methods and show that semantic pairs consistently improve downstream transfer performance, with contrastive learning benefiting most strongly.
    \item     We conduct extensive ablation studies to evaluate the sensitivity of semantic pairing to various data transformations, architectural choices, dataset sizes, and occlusion robustness.

\end{itemize}

\section{Related Work}
SSL has emerged as a powerful paradigm for learning representations from unlabeled data. SSL approaches are divided into two broad categories: contrastive and non-contrastive learning. These approaches often use pretext tasks to learn useful representations from unlabeled data. This section provides a brief overview of some of these approaches and benchmark datasets, such as ImageNet or CIFAR-10, that are used with SSL. We encourage readers to refer to the respective papers for more details.

\textbf{Representation Learning:}
Various theoretical studies demonstrate how models learn visual representations from unlabeled data from a mathematical perspective, explaining the underlying mechanisms of contrastive and representation learning objectives \citep{haochen2021provable,johnson2022contrastive,zhai2025contextures,pmlr-v267-reizinger25a,alshammari2025a}. Specifically, \cite{haochen2021provable} replace conditional independence assumptions with population graphs to prove neural networks achieve linear separability through spectral graph decomposition, whereas \cite{johnson2022contrastive} show that minimizing InfoNCE acts as Kernel Principal Component Analysis to discover an optimal linear basis for view-invariant tasks. To expand beyond standard contrastive bounds, \cite{zhai2025contextures} prove that pretraining approximates singular functions of association operators (establishing hard mathematical limits on parameter scaling), while \cite{pmlr-v267-reizinger25a} propose Singular Identifiability Theory (SITh) to model realistic data-generating processes and explain model convergence toward shared Platonic representations. Providing a universal taxonomy, \cite{alshammari2025a} introduce the Information Contrastive Learning (I-Con) framework to unify over twenty distinct self-supervised, supervised, and clustering algorithms into a single neighborhood KL-divergence objective. While these works provide mathematical justifications for standard single-instance transformations, our study shifts the focus to investigating the impact of semantic pairs on self-supervised representation learning. Specifically, we conduct a controlled empirical investigation using a matched-scale ImageNet-1K subset to uniquely isolate semantic positive pairs from standard same-instance augmentations. By doing so, we empirically prove that semantic pairs induce stronger representation invariance and significantly enhance downstream generalization on unseen datasets.

\textbf{Dataset:}
Datasets like STL-10 \citep{coates2011analysis}, CIFAR \citep{krizhevsky2009learning}, Tiny-ImageNet \citep{le2015tinyimagenet}, ImageNet-100, and ImageNet-1K \citep{russakovsky2015imagenet}, all organise images into labelled classes, which is a fundamental structure. However, they vary significantly in size and complexity, making them valuable benchmark datasets for various machine learning tasks. Notably, none of these datasets explicitly contains semantic pairs, requiring SSL models to identify these relationships directly from the data during training. In this study, we create a dataset containing curated semantic pairs, which reduces the issues of mapping inaccurate semantic pairs during model training (i.e., false neighbors). Thus, this allows us to focus on studying the effect of semantic pairs in the SSL context. 

\textbf{Contrastive Learning:} Instance discrimination methods, such as SimCLR, MoCo, and PIRL \citep{chen2020simple,he2020momentum,chen2020improved,misra2020self} employ a similar idea. They attract the positive pairs together and push the negative pairs apart in the embedding space, albeit through a different mechanism.  SimCLR \citep{chen2020simple} employs an end-to-end approach where a large batch size is used for the negative examples and both encoders’ parameters in the Siamese network are updated together. PIRL \citep{misra2020self} uses a memory bank for negative examples, and both encoders’ parameters are updated together. MoCo \citep{chen2020improved,he2020momentum} employs a momentum contrastive approach, where the query encoder is updated during backpropagation and, in turn, updates the key encoder. Negative examples are stored in a separate dictionary, allowing for the use of large batch sizes.

\textbf{Non-Contrastive Learning:}
Non-contrastive approaches use only positive pairs to learn visual representations employing different methods to avoid representation collapse. The first category is clustering-based methods, where samples with similar features are assigned to the same cluster. DeepCluster \citep{caron2018deep} uses pseudo-labels from the previous iteration, which makes it computationally expensive and hard to scale. SWAV \citep{caron2020unsupervised} addresses this issue by using online clustering, though it requires determining the correct number of prototypes. The second category involves knowledge distillation. Methods like BYOL \citep{grill2020bootstrap} and SimSiam \citep{chen2021exploring} draw on knowledge distillation techniques, where a Siamese network comprises an online encoder and a target encoder. The target network parameters are not updated during backpropagation. Instead, only the online network parameters are updated while being encouraged to predict the representation of the target network. Despite their promising results, the mechanisms by which these methods avoid collapse are not yet fully understood. Inspired by BYOL, Self-distillation with no labels (DINO) \citep{caron2021emerging} uses centering and sharpening, along with a different backbone (ViT), which allows it to outperform other self-supervised methods while being more computationally efficient. 
Another approach, Bag of visual words \citep{gidaris2020learning,gidaris2021obow}, utilises a teacher-student framework inspired by natural language processing (NLP) to avoid representation collapse. The student network predicts a histogram of the features for augmented images, similar to the teacher network’s histogram. The last category is information maximisation. Methods like Barlow twins \citep{zbontar2021barlow} and VICReg \citep{bardes2021vicreg} do not require negative examples, stop gradient or clustering. Instead, they use regularisation to avoid representation collapse. The objective function of these methods tries to eliminate the redundant information in the embeddings by making the correlation of the embedding vectors closer to the identity matrix. While these methods show promising results, they have limitations, such as the representation learning being sensitive to regularisation and decreased effectiveness if certain statistical properties are missing in the data.

\textbf{Enhanced Instance Discrimination:} Efforts to enhance contrastive learning have significantly focused on refining negative examples to improve representation quality. For instance, \citep{kalantidis2020hard} and \citep{robinson2020contrastive} emphasized mining "hard" negative samples close to the positive anchor, while \citep{wu2020conditional} proposed a percentile-based strategy for negative sampling. Another approach by \citep{chuang2020debiased} involved assigning weights to positive and negative terms to mitigate the impact of irrelevant negatives. \citep{dwibedi2021little,koohpayegani2021mean, huynh2022boosting,10296635,alkhalefi2023semantic,zeng2024contrastive} enhance representation learning by utilising semantic pairs to increase data diversity and learn invariant representations to intra-class variation.

Although the aforementioned approaches utilise semantic pairs to enhance representation learning in instance discrimination, accurately discerning the specific contribution of semantic pairs to the observed improvements in model performance requires careful and rigorous analysis. A potential confounding factor arises from the integration of semantic pairs with either augmented pairs or false neighbours in existing approaches, complicating the isolation of the unique impact of semantic pairs on self-supervised learning. In addition, the mentioned approaches employed semantic pairs with either a contrastive approach, such as NNCLR \citep{dwibedi2021little}, or a non-contrastive approach, such as MSF \citep{koohpayegani2021mean}, without providing clear justification for their choice of approach with semantic pairs. Therefore, this study aims to conduct a rigorous investigation to precisely determine the impact of semantic pairs on the performance and generalization capabilities of the SSL approaches. Additionally, it seeks to identify the most effective approach for leveraging semantic pairs to learn invariant representations, thereby enhancing the model’s generalization on unseen datasets. To achieve this, a curated dataset containing semantic pairs is constructed to eliminate confounding factors present in previous approaches. Subsequently, a comparative analysis is performed by training one model on this meticulously curated semantic pairs dataset and another on standard augmented pairs.

\section{Methodology}
This study investigates whether incorporating semantically paired images can enhance the generalization capability of SSL models. To assess this, we compare two pretrained models: one using standard augmented image pairs (our baseline) and another pre-trained on semantically related pairs, both derived from the ImageNet-1K dataset \citep{russakovsky2015imagenet}. As illustrated in Figure 3, our comparative methodology comprises four key stages: (1) dataset construction, (2) image transformations, (3) model training, and (4) performance evaluation.

\begin{figure*}[h]
\begin{center}
\includegraphics[width=1.05\textwidth]{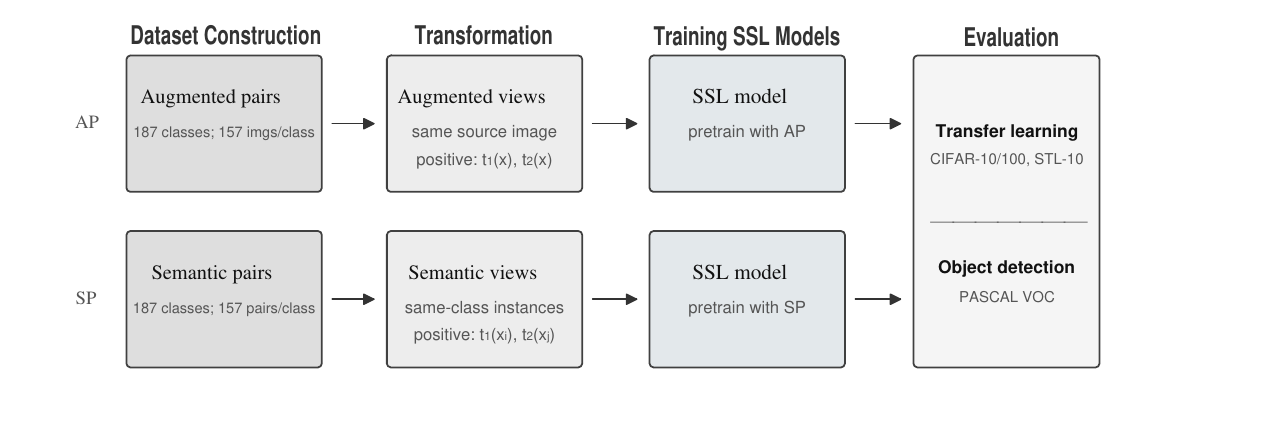}

\end{center}
   
   \caption{Controlled experimental framework for comparing augmented and semantic positive pairs. The augmented-pair baseline uses two stochastic views of the same instance, whereas the semantic-pair dataset uses two different instances from the same class. Both branches are strictly matched in class composition, training-pair count, backbone, training schedule, and transformation pipeline. Downstream performance is compared using transfer learning and object detection.}
\label{fig:methodology}
\end{figure*}

\subsection{Dataset Construction}
We developed two distinct datasets derived from the ImageNet-1K training split: one comprising augmented pairs (serving as the baseline) and the other consisting of semantic pairs. Both datasets were balanced with respect to the number of classes and images per class to ensure a fair comparison. 

\subsubsection{Semantic Pairs Dataset}
We construct a manually curated dataset of semantic pairs, comprising 187 classes, each containing 157 pairs, resulting in a total of 29,359 semantic pairs (see Appendix A.4).
Each pair is manually annotated to ensure precise semantic alignment, mitigating potential misalignments inherent in automated matching methods. For every semantic pair, we identify instances that exhibit semantic correspondence (i.e., the same object) while varying in contextual attributes (See Appendix A (Figures 6 and 7) for illustrated examples).   Since the primary objective of this study is to evaluate model generalization on unseen data, we adopted a structured strategy for class selection rather than sampling random categories from ImageNet-1K. Specifically, we prioritized classes that exhibit semantic overlap with downstream evaluation datasets such as STL-10 and CIFAR. For example, STL-10 consists of ten categories (airplane, bird, car, cat, deer, dog, horse, monkey, ship, and truck). Corresponding categories were identified within ImageNet and integrated into the semantic pairs dataset. This facilitates the evaluation of which model learns more effective representations. 
For instance, if a model is pretrained on two classes (e.g., dog and cat) but the downstream task involves distinct classes (e.g., car and airplane), both models may exhibit comparably low performance due to the cross-category shift between the pretraining and downstream datasets.

\vspace{-3 mm}
\subsubsection{Augmented Pairs Dataset (baseline)}
\vspace{-2 mm}
The augmented pairs dataset comprises 187 classes, each containing 157 images, resulting in a total of 29,359 images, which maintains consistency with the Semantic pair dataset in both class composition and scale. 
\vspace{-4 mm}
\subsection{Transformation}
\vspace{-2 mm}
During model training, a standard stochastic transformation pipeline is applied, including random cropping, rotation, random horizontal flipping, color jittering, and Gaussian blur, consistent with the approaches included in this study. For the augmented dataset (29,359 images), this process generates synthetic pairs by creating multiple augmented views of the same instance, resulting in 29,359 pairs. In contrast, the semantic dataset consists of predefined pairs, so transformations are applied individually to each image within these pairs without generating additional synthetic pairs. Figure 1 illustrates examples of both transformation approaches, where Figure 1(a–b) presents data transformations for augmented and semantic pairs, respectively.
\vspace{-2 mm}
\subsection{Training SSL Models}
\vspace{-2 mm}
In this step, we train multiple SOTA self-supervised approaches on both the augmented pair dataset and the semantic pair dataset. First, we compare models trained on augmented pairs against those trained on semantic pairs to evaluate whether semantic pairs help the model generalize better to unseen data. Next, we measure the relative improvement each SSL method achieves when using semantic pairs versus augmented pairs, quantifying the extent to which each model benefits from semantic relationships.
This comparison reveals which SSL approach leverages semantic pairs most effectively to learn useful representations.
\vspace{-2 mm}
\subsection{Evaluation}
\vspace{-2 mm}
After creating both datasets, we eliminate the confounder present in previous approaches, where semantic pairs were combined with either positive pairs or false neighbors. This design enables a direct and fair assessment of their influence on SSL. As is well established, SSL methods are typically evaluated using downstream tasks to assess the quality of learned representations. In this study, model performance is evaluated using two primary downstream tasks: transfer learning and object detection. These tasks assess model generalization on unseen datasets, thereby demonstrating the ability to learn discriminative and transferable features that enable accurate predictions on new data. To ensure rigorous and unbiased evaluations, our experimental protocol is structured into two distinct analytical phases: intra-method self-comparison and inter-method cross-comparison.

\subsubsection{Intra-Method Evaluation: Self-Comparison}
To evaluate the direct impact of incorporating semantic pairs within any individual framework, we compare a model pretrained on semantic pairs directly against its respective baseline, which is pretrained on standard augmented pairs. To ensure a fair self-comparison, we strictly replicated the method-specific hyperparameter configurations reported in each approach's original publication, including optimizer configurations, learning rate schedules, and warm-up epochs. Because these method-specific hyperparameters were held strictly constant between the baseline and the model pretrained on semantic pairs, any observed performance improvement ($\Delta$) can be attributed solely and definitively to the incorporation of semantic pairs.

\subsubsection{Inter-Method Evaluation: Cross-Approach Comparison}
To determine which SSL framework most effectively utilizes semantic information, we compare the achieved performance improvements ($\Delta$) across the different approaches. A valid cross-comparison requires controlling for structural and scaling confounders. Without this control, evaluating models with mismatched configurations, such as comparing SimCLR trained with a batch size of 256 against BYOL at 2048 or comparing a ResNet-50 backbone against a Vision Transformer (ViT), would yield invalid conclusions. Consequently, all approaches were pretrained using strictly identical foundational variables: backbone architecture, batch size, number of epochs, and data transformation pipelines. Because these structural parameters were rigorously standardized across all frameworks, the approach exhibiting the largest performance difference ($\Delta$) between its baseline and the model pretrained on semantic pairs is conclusively demonstrated to leverage semantic pairs most effectively for learning high-quality representations.

\vspace{-2 mm}
\section{Experiments}
\vspace{-2 mm}
\textbf{Training Setup.} We pre-trained several SOTA approaches on two distinct datasets: the Augmented Pairs Dataset and the Semantic Pairs Dataset. The experiments utilized a ResNet50 backbone, with a batch size of 256, and a training duration of 200 epochs. All remaining hyperparameters and configurations corresponding to each method were adopted from the original publications.

\textbf{Evaluation.}
Following the training phase, two downstream tasks are employed for evaluation: transfer learning and object detection. In transfer learning, the backbone is frozen, and a classification head is appended to it. Transfer learning experiments are conducted on three benchmark datasets: STL-10 \citep{coates2011analysis}, CIFAR-10, and CIFAR-100 \citep{krizhevsky2009learning}. In addition, object detection fine-tuning is performed on PASCAL VOC \citep{everingham2010pascal}, where the backbone and detection head are jointly trained. These evaluations aim to determine whether the incorporation of semantic information enhances the generalization capability of SSL approaches and, if so, which approach performs best.

\textbf{SOTA Methods on Semantic and Augmented Pairs.} In our experimental evaluation, we applied four SOTA methods across two datasets: one comprising semantically related pairs and another consisting of augmented pairs, yielding a total of eight distinct trained models for comparison. The selection included both contrastive and non-contrastive representation learning approaches. While individual methods within each approach differ in their implementation specifics, they generally operate based on a shared underlying principle, resulting in comparable overall learning behavior. Accordingly, our analysis is conducted at the approach level rather than focusing on individual methods. Specifically, SimCLR \citep{chen2020simple} was selected as a representative of contrastive learning. For non-contrastive learning, three representative methods were chosen to capture different subfamilies: VICReg \citep{bardes2021vicreg}, representing the information maximization family (e.g., VICReg and Barlow Twins); BYOL \citep{grill2020bootstrap}, exemplifying self-distillation frameworks that employ architectural constraints to prevent representational collapse (e.g., BYOL and SimSiam); and DINO \citep{caron2021emerging}, a distinct self-distillation approach incorporating centering and sharpening mechanisms to mitigate collapse. The comparative transfer learning performance of these approaches is presented in Figure 4.

\subsection{Transfer Learning}
Figure 4 shows that models pretrained on semantic pairs consistently outperform those pretrained on augmented pairs when evaluated on unseen data. This suggests that the semantic pairs dataset provides richer semantic information, thereby enhancing the model’s ability to generalize to unseen datasets. These results align with established findings in transfer learning, which demonstrate that models pretrained on semantically rich data transfer more effectively across datasets and achieve superior generalisation \citep{yosinski2014transferable}.
Figure 4 also shows that the contrastive learning approach consistently outperforms competing approaches in learning useful representations from semantic pairs.   Among the evaluated methods, SimCLR achieves the highest performance across all downstream datasets.
For instance, SimCLR (SP), which is pretrained on a semantic pairs dataset, achieves a 3.83\% improvement in transfer learning to the STL-10 dataset compared to its counterpart, augmentation-based SimCLR (AP). 

\begin{figure*}[h]
\begin{center}
\includegraphics[width=1\textwidth]{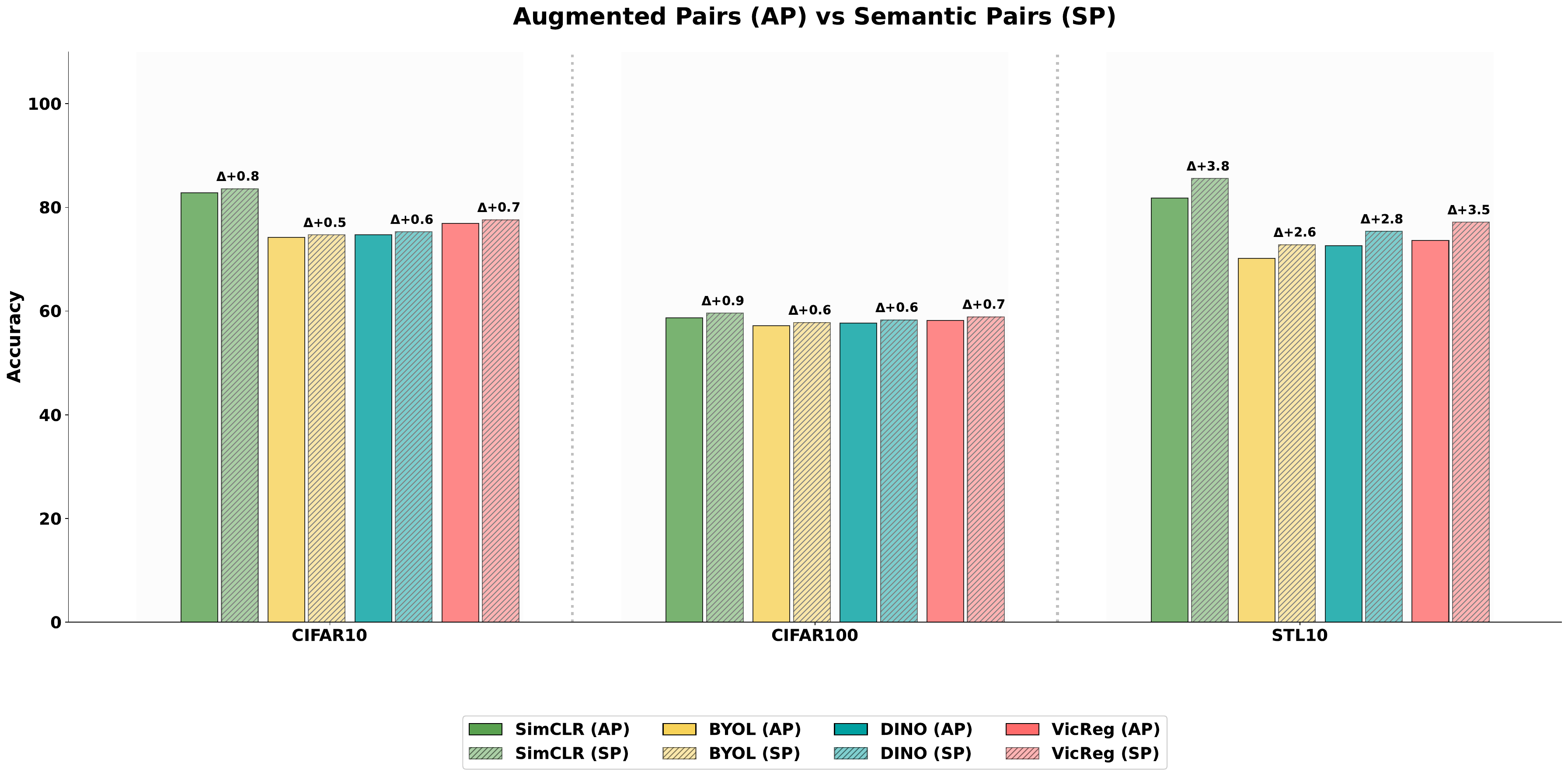}

\end{center}
   
   \caption{ Transfer learning performance of SOTA approaches pretrained on Semantic Pairs (SP, striped) and Augmented Pairs (AP, solid), evaluated on CIFAR10, CIFAR100, and STL10. (SP) Pretraining consistently improves downstream accuracy over(AP). ($\Delta$) values denote (SP-AP) differences.}
\label{fig:results}
\end{figure*}

To evaluate which approach learns semantic features more effectively from semantic pairs, we compare the performance differences ($\Delta$) between approaches rather than their overall accuracy values. For each approach, such as SimCLR, the performance difference ($\Delta$) is obtained by subtracting the performance of the model pretrained on augmented pairs from that of the model pretrained on semantic pairs. This provides a more accurate indication of which approach learns semantic features more effectively from the input data.  For example, consider the case where SimCLR (SP), trained on semantic pairs, achieves 85.59\% accuracy when transferred to the STL-10 dataset, while VICReg (SP), trained on semantic pairs, reports 87\%.  A direct comparison of these results does not necessarily imply that VICReg learns semantic features more effectively; the difference may be due to VICReg’s method performing better than SimCLR, rather than being attributed to semantic pairs. To address this issue, we compare the relative improvements of each method. For SimCLR, accuracy increases from 81.76\% on the augmented pairs dataset to 85.59\% on the semantic pairs dataset, corresponding to an improvement of +3.83\%. In contrast, VICReg increases from 85.5\% on augmented pairs to 87.0\% on semantic pairs, yielding a smaller improvement of +1.5\%. Although VICReg (SP) achieves higher absolute accuracy, SimCLR demonstrates a larger improvement when trained on semantic pairs, indicating that it benefits more from semantic supervision. In other words, the VICReg might be better in absolute accuracy, but SimCLR is better in learning from semantic pairs because the ($\Delta$) between SimCLR (AP) and SimCLR (SP) is greater than the ($\Delta$)  of VICReg.  This comparison enables researchers to identify the most suitable SSL framework for leveraging semantic pairs, rather than selecting a framework arbitrarily. Additionally, it motivates further investigation into why contrastive learning methods (e.g., SimCLR) capture semantic features more effectively than non-contrastive approaches (e.g., VICReg) from semantic pairs. Understanding this dynamic will help optimize the exploitation of semantic pairs within non-contrastive models, thereby enhancing representation learning and generalization across SSL approaches.

\begin{table}[h]
\caption{Comparison of Transfer Learning Performance on STL-10 after 800 epochs of pre-training for SimCLR with Augmented Pairs (AP) and Semantic Pairs (SP).}
\label{tab:table1}
\centering
\begin{tabular}{l c c c}
\hline
\textbf{Approach} & \textbf{200 Epochs} & \textbf{400 Epochs} & \textbf{800 Epochs} \\
\hline\hline
SimCLR (SP) & 85.59\% & 86.50\% & 86.56\% \\
SimCLR (AP) & 81.76\% & 82.23\% & 82.41\% \\
\hline
\textbf{Improvement} & \textcolor{my_green}{+3.83\%} & \textcolor{my_green}{+4.27\%} & \textcolor{my_green}{+3.75\%} \\
\hline
\end{tabular}
\end{table}

SimCLR is selected for further analysis, as it demonstrates superior performance among the evaluated approaches leveraging semantic pairs. To assess whether semantic pairs consistently enhance model generalisation, the training schedule is extended to 800 epochs. As presented in Table 1, SimCLR (SP) consistently outperforms SimCLR (AP) across all epochs. Both models converge after 800 epochs; however, SimCLR (SP) maintains a performance margin of 3.75\% over SimCLR (AP).

\subsection{Object Detection}

This evaluation assesses the generalization capability of representations learned by semantic pairs, with a particular focus on their effectiveness beyond standard classification tasks. The model was fine-tuned for object detection using the PASCAL VOC dataset. Following the experimental protocol of MoCo-v2\citep{chen2020improved}, fine-tuning was conducted on the combined VOC07+12 trainval set employing a Faster R-CNN architecture with a R50-C4 backbone. Evaluation was performed on the VOC07 test set. The fine-tuning process comprised 24K iterations, corresponding to approximately 23 epochs. 
Table 2 presents a comparative analysis of performance between SimCLR pretrained on semantic pairs and SimCLR pretrained on augmented pairs. The evaluation metrics considered include $AP_{50}$, $AP$, and $AP_{75}$ . The results indicate that SimCLR (SP) pretrained on semantic pairs consistently outperforms the model pretrained on augmented pairs across all metrics, thereby suggesting that pretraining on semantic pairs substantially improves both object localization and classification performance.

\begin{table}[h]
\caption{\textcolor{black}{Results (Average Precision) for PASCAL VOC object detection using Faster R-CNN with ResNet50-C4.}}

\centering
\begin{tabular}{c c c c }

\hline
Approach  & $AP_{50}$ & $AP$ & $AP_{75}$\\
\hline\hline
SimCLR (SP)& 75.02\%& 50.30\% & 55.22\% \\
\hline
SimCLR (AP) &73.82\%&48.9\%&53.72\% \\
\hline

\hline

\end{tabular}

\label{tab:table3}
\end{table}

SimCLR (SP) achieves the highest gain in $AP_{75}$, with 55.22 \% versus 53.72 \%. This metric, which requires a (75\%) overlap between predicted and true bounding boxes, highlights the model’s ability to perform precise localisation. The model also improves by 1.2 \% in $AP_{50}$, and 1.4 \% in overall $AP$. These results indicate that pretraining on semantic pairs exposes the model to objects across diverse contexts, enabling it to focus more effectively on object features rather than background cues. This exposure fosters greater invariance to background variations, thereby improving both recognition robustness and localization precision \citep{Zhao_2022_CVPR}. Consequently, semantic pair pretraining enhances representation quality and strengthens the model’s ability to generalise to downstream object detection tasks more effectively than transformation-based augmentation alone.

\section{Ablation Studies} \label{sec:ablation}
In the ablation studies, a series of experiments is conducted to further investigate the impact of semantic pairs on SSL. First, data transformation variants are removed from the training pipeline, and models are trained on both datasets to determine if semantic pairs alone can impose invariance and help SSL models learn more robust semantic feature representations. Next, the ResNet-50 backbone is replaced with a Vision Transformer (ViT), followed by transfer learning and fine-tuning, to assess whether the generalization improvement from semantic pairs is consistent across different architectures and downstream tasks. Subsequently, a random crop test is introduced to simulate occlusion scenarios, presenting only partial object views instead of full ones obtained through center cropping. This step examines whether incorporating semantic information improves the model’s robustness to occlusion. Finally, models are trained on datasets of varying sizes to investigate how performance responds to different quantities of semantic pairs. This design helps analyze whether smaller datasets limit generalization and lead to learning patterns similar to models pretrained with augmented pairs, offering insights into semantic pair sensitivity and data efficiency. 

\subsection{Evaluating Model Performance with Removed Transformations}
 In these experiments, we systematically ablate variant transformation techniques from the pipeline and subsequently retrain the models on both datasets. By doing this, we can see if the semantic pairs impose invariance by placing the object in different contexts, which helps the model learn a generalized representation for the input object. 

\begin{figure*}[h]
\begin{center}
\includegraphics[width= .85\textwidth]{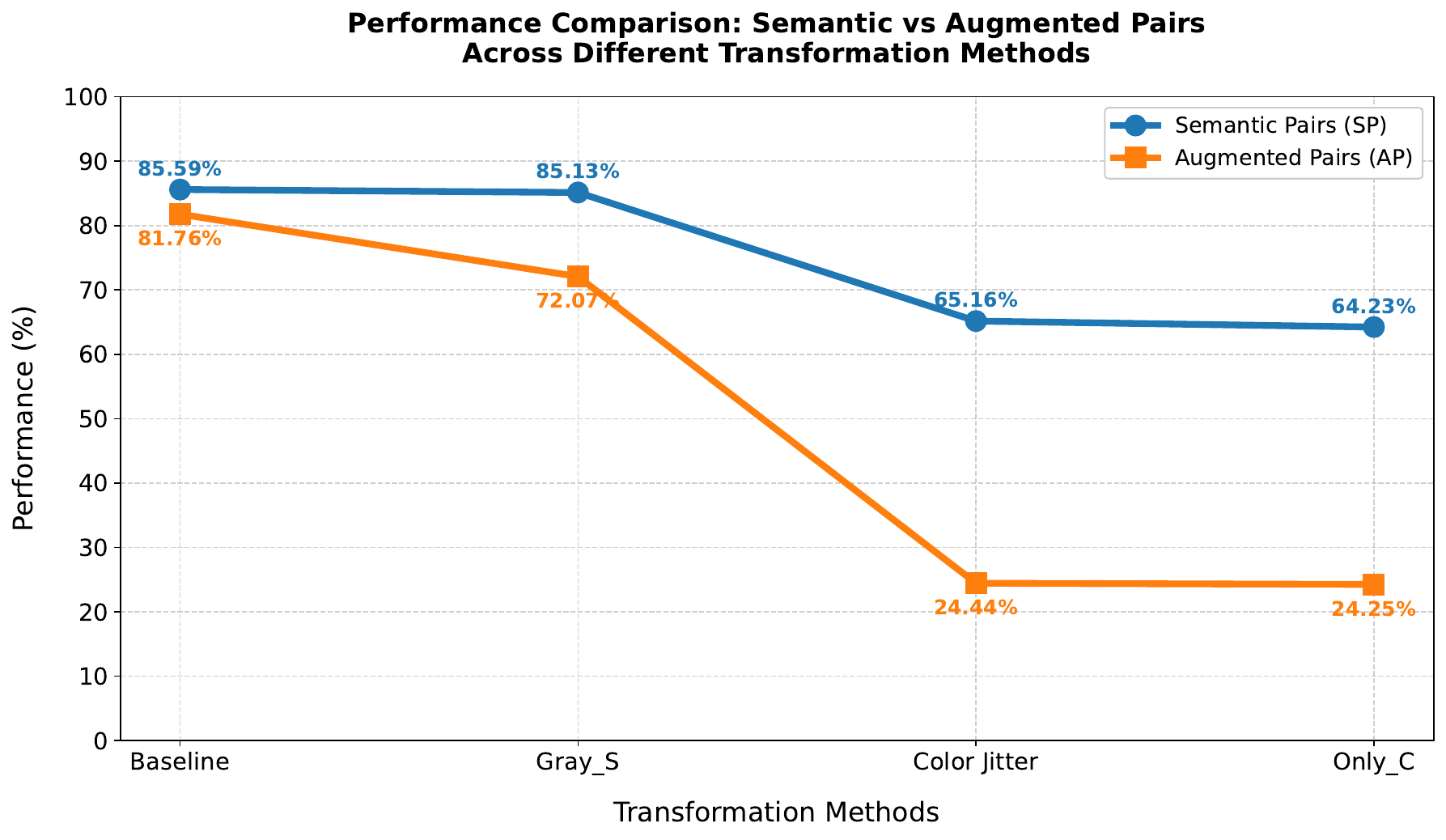}

\end{center}
   
   \caption{Performance comparison of SimCLR (SP) and SimCLR (AP) pre-trained for 200 epochs with transformation ablations: (1) Gray$\_$s (grayscale removed), (2) color$\_$jitter (both color jitter and grayscale removed), and (3) only$\_$c (only random crop retained).}
\label{fig:transsformation}
\end{figure*}

As illustrated in Figure 5, the model pretrained on the augmented dataset exhibits a substantial decline in performance, reaching 24.25\%, when all transformation techniques are removed from the pipeline except random cropping (Only$\_$C). This degradation occurs because the model becomes increasingly sensitive to illumination, background colour, and superficial object features such as logos and decorations. Conversely, the model pretrained on the semantic pairs dataset demonstrates a notably higher robustness, achieving 64.23\% under the same conditions. This result empirically indicates that semantic pairs induce invariances during representation learning, allowing the model to learn a more invariant representation for the input data. In essence, semantic pairs facilitate more effective representation learning by exposing the model to diverse contextual variations, thereby enhancing its ability to generalize beyond predefined transformations.

\subsection{Generalization Across Architectures and Tasks}

In this experiment, we conduct further assessment for the robustness of the semantic pairs in supporting SSL generalization. This study extends the analysis to encompass both an alternative backbone architecture and a distinct downstream task. Specifically, the ViT-S/8 model is pretrained on both datasets, followed by transfer learning via fine-tuning to assess downstream performance. Among the methods examined in this study, only the DINO framework natively supports the ViT architecture. Accordingly, DINO is trained with the ViT backbone on both datasets. The training protocol utilizes a batch size of 256 and 300 epochs. In this experiment, we employed only two global views per sample, while all remaining hyperparameters adhere to those defined in the original DINO setup. The results presented in Table 3 indicate that DINO (SP) with semantic pairings consistently outperforms DINO (AP) with augmented pairings across all evaluations. These outcomes provide robust evidence that semantic pairs enhance the generalization capability of SSL  irrespective of the architectural framework or downstream task considered.

\begin{table}[h]
\caption{\textcolor{black}{Comparison of DINO (ViT-S/8) fine-tuning results after 300 epochs of pretraining on semantic pairs (SP) and augmented pairs (AP).}}

\centering
\begin{tabular}{c c c c }

\hline
Approach  & \textbf{CIFAR-10} & \textbf{CIFAR-100} & \textbf{STL-10}\\
\hline\hline
DINO (SP)& 81.8\%& 65.3\% & 82.1\% \\
\hline
DINO (AP) &81.1\%&64.5\%&79.2\% \\
\hline

\hline

\end{tabular}

\label{tab:table4}
\end{table}

\subsection{Centre and Random Crop Test}
 In this experiment, both models are evaluated using randomly cropped images from the STL-10 dataset, rather than the centred object regions typically used. This testing approach presents only a portion of the object, positioned variably within the image, thereby increasing the difficulty of accurately predicting the correct label. The experiment is designed to simulate real-world conditions, where objects are not always centrally located or fully visible within an image. Superior performance under these conditions indicates that the model exhibits stronger occlusion invariance, demonstrating its ability to infer the correct semantic label when objects only partially appear.

\begin{table}[h]
\caption{Comparison of Transfer Learning Performance on STL-10 after 200 epochs of pre-training for SimCLR with Augmented Pairs (AP) and Semantic Pairs (SP).}
\label{tab:table5}
\centering
\begin{tabular}{l c c c}
\hline
\textbf{Approach} &  \textbf{Center Crop} & \textbf{Random Crop} \\
\hline\hline
SimCLR (SP) & 85.59\% &  79.29\% \\
SimCLR (AP) & 81.76\% & 74.26\% \\
\hline

\end{tabular}
\end{table}

As shown in Table 4, the performance of the model pretrained with augmented pairs SimCLR (AP) decreases by 7.5\%, while that of the semantic pairs SimCLR (SP) declines by 6.3\%. This indicates that incorporating semantic pairs promotes occlusion invariance in self-supervised learning, enabling the model to perform more effectively when only a portion of the image is visible, thereby enhancing its generalization capability.

\subsubsection{Impact of Training Set Size on Generalisation}
In this experiment, we evaluate the models by training them on datasets of varying sizes (50, 100, and 157 images per class) for 200 epochs. After training, we apply transfer learning to the STL-10 dataset to assess its impact on generalisation performance. This design enables a systematic investigation of the relationship between the semantic pairs dataset size and the models’ ability to generalise. 

\begin{table}[h]
\caption{Comparison of Transfer Learning Performance on STL-10 after 200 epochs of pre-training for SimCLR with Augmented Pairs (AP) and Semantic Pairs (SP).}
\label{tab:table6}
\centering
\begin{tabular}{l c c c}
\hline
\textbf{Approach} & \textbf{50 Images} & \textbf{100 Images} & \textbf{157 Images} \\
\hline\hline
SimCLR (SP) & 81.68\% & 83.01\% & 85.59\% \\
SimCLR (AP) & 77.48\% & 78.97\% & 81.76\% \\
\hline
\textbf{Improvement} & \textcolor{my_green}{+4.20\%} & \textcolor{my_green}{+4.04\%} & \textcolor{my_green}{+3.83\%} \\
\hline
\end{tabular}
\end{table}

While Table 5 indicates that SimCLR (SP), pretrained on semantic pairs, consistently outperforms SimCLR (AP), which relies solely on the transformation pipeline, across different dataset sizes, an important trend can be observed. Specifically, when the number of training examples is increased to 157 per class, the performance gap between the two models decreases to 3.83\%. In contrast, when the number of examples is reduced to 50 per class, the gap widens to 4.20\%. This widening occurs because when the number of samples is reduced to 50 per class, the traditional transformation pipeline fails to capture sufficient intra-class variability, leading to weaker invariant representations and, consequently, reduced generalization to unseen data. In other words, increasing the number of images in the augmented pairs dataset allows the model to be exposed to diverse data, which allows it to be more invariant and improve its generalization closer to semantic pairs on the unseen dataset. In contrast, semantic pairs, which comprise images that present the same object in varying contexts, inherently provide greater data diversity. This richer diversity enables the model to become more invariant to intra-class variations, leading to more robust and generalizable results, even in scenarios with limited training data. This is evident in Table 5, where training SimCLR (SP) on 50 images per class yields performance comparable to that of SimCLR (AP) trained on 157 images per class. This observation aligns with the hypothesis of this study, which posits that leveraging semantic positive pairs can enhance representation learning by exposing models to meaningful variations beyond the limited variations of the transformation pipeline.


\section{Conclusion}

This study examined the role of semantic positive pairs in self-supervised visual representation learning. To isolate their effect from standard same-instance augmented positives and false neighbors, we constructed two matched ImageNet-1K subsets: an augmented-pair baseline and a manually curated semantic-pair dataset with the same class composition and training-pair count. This controlled design enabled a direct comparison between models trained with augmented positive pairs and those trained with semantic positive pairs.

Across transfer learning and object detection evaluations, models pretrained with semantic pairs consistently outperformed their augmented-pair counterparts. These results suggest that semantic pairs provide useful invariance beyond that induced by conventional transformation pipelines, helping models learn representations that transfer more effectively to unseen datasets. The ablation studies further support this interpretation: semantic-pair pretraining remained more robust when transformations were removed, improved performance under random-crop evaluation, generalized across a ViT-based DINO setting, and retained advantages across different dataset sizes.

Among the evaluated SSL methods, contrastive learning benefited most strongly from semantic pairs, with SimCLR showing the largest relative improvement over its augmented-pair baseline. This finding suggests that the interaction between semantic positive pairs and the training objective is important: not all SSL frameworks exploit semantic relationships equally effectively. Understanding why contrastive objectives appear to benefit more from semantic-pair information remains an important direction for future work.

Overall, the results indicate that semantic positive pairs can serve as a valuable complement to transformation-based augmentation in self-supervised learning. By exposing models to same-class objects across diverse contexts, semantic pairs encourage representations that are less dependent on nuisance factors such as background, illumination, occlusion, and superficial visual details. Future work should investigate larger-scale semantic-pair construction, automated pair selection, stronger controls for the number of unique source images, and objective functions designed specifically to exploit semantic positive relationships.

\section*{Future Work}

While our current scope demonstrates the efficacy of semantic pairs, future work will explicitly evaluate cross-category generalization by transferring learned representations to completely unseen classes. Specifically, we will investigate scenarios representing a "severe cross-category shift," distinct from standard discrepancies in data distributions. Furthermore, our empirical results indicate that contrastive learning frameworks benefit more significantly from semantic pairs than non-contrastive methods. Consequently, we will investigate the underlying mechanisms driving this difference to understand why contrastive objectives are better equipped to leverage semantic pairs to learn invariant representations that enable the model to generalize to new datasets.

\section*{Acknowledgments}
We would like to thank the University of Aberdeen’s High Performance Computing facility for enabling this work.

\bibliography{main}
\bibliographystyle{tmlr}

\appendix
\section{}
\label{app:documentation}

\textbf{Dataset Details}

We introduce a dataset of curated semantic pairs derived from ImageNet-1K, comprising \textbf{187} carefully selected classes (from the original 1,000) designed to advance representation learning through semantically meaningful instance pairs. Our class selection was strategically aligned with standard benchmarks, such as STL-10, to ensure optimal transfer learning performance.

\subsection{Key Features}
\begin{itemize}
    \item Maintains original ImageNet-1K image resolutions
    \item Each instance is \textbf{paired semantically} with another instance from the same class
    \item These curated pairs serve as \textbf{ground-truth supervision} for learning tasks
    \item The pairing strategy is designed to:
    \begin{enumerate}
        \item Capture meaningful \textbf{semantic similarities} (e.g., invariant features across different object poses or lighting conditions)
        \item Minimize attention to \textbf{irrelevant variations} (e.g., background clutter or texture biases)
    \end{enumerate}
\end{itemize}

\subsection{Dataset Composition}
\begin{itemize}
    \item \textbf{187 classes} covering diverse categories (animals, objects, landscapes, etc.)
    \item \textbf{157 semantic pairs} per class (29,359 pairs total)
    \item Classes intentionally selected to mirror standard benchmark datasets
\end{itemize}

\subsection{Construction Details}
\begin{itemize}
    \item Required \textbf{6 months} of full-time manual curation (8 hours/day).
    \item Ensures \textbf{high-quality labels} through human verification.

\end{itemize}

\subsection{Dataset Classes}

This subsection illustrates the 187 classes selected from ImageNet-1K.  

\begin{itemize}[leftmargin=*,nosep]
    \item \textbf{Animals}:\\
    African\_grey, albatross, American\_alligator, American\_coot, American\_egret, American\_lobster, Afghan\_hound, Arabian\_camel, bald\_eagle, basset, beagle, bee\_eater, bittern, black\_grouse, black\_stork, black\_swan, black\_widow, Blenheim\_spaniel, brambling, brown\_bear, bullfrog, bulbul, bustard, centipede, chickadee, Chihuahua, chimpanzee, cock, cockroach, cougar, crane, drake, dowitcher, Egyptian\_cat, European\_fire\_salamander, flamingo, gazelle, goldfinch, goldfish, goose, great\_grey\_owl, hen, hornbill, house\_finch, hummingbird, indigo\_bunting, jacamar, Japanese\_spaniel, jay, king\_penguin, kite, lemon\_shark, limpkin, little\_blue\_heron, lion, macaw, magpie, Maltese\_dog, monarch, oystercatcher, peacock, pelican, Pekinese, ptarmigan, quail, red-breasted\_merganser, red-backed\_sandpiper, redshank, Rhodesian\_ridgeback, robin, ruffed\_grouse, Shih-Tzu, snail, spoonbill, sulphur\_butterfly, sulphur-crested\_cockatoo, tabby, tadpole, tailed\_frog, tarantula, tiger\_shark, toucan, toy\_terrier, vulture, water\_ouzel, white\_stork, birdhouse.

    \item \textbf{Plants \& Fungi}:\\
    banana, bell\_pepper, corn, daisy, Granny\_Smith, lemon, mushroom, orange, pomegranate, rapeseed, strawberry, yellow\_lady's\_slipper

    \item \textbf{Vehicles}:\\
    aircraft\_carrier, airliner, airship, ambulance, amphibian, balloon, bullet\_train, cab, convertible, electric\_locomotive, fire\_engine, forklift, freight\_car, garbage\_truck, go-kart, golfcart, jeep, limousine, minivan, Model\_T, moving\_van, passenger\_car, pickup, police\_van, racer, recreational\_vehicle, schooner, snowplow, space\_shuttle, sports\_car, streetcar, trailer\_truck, tractor, tow\_truck, thresher, trolleybus, unicycle, yawl

    \item \textbf{Architectural \& Natural Landmarks}:\\
    alp, cliff, geyser, lakeside, promontory, sandbar, seashore, triumph\_arch, valley, viaduct, volcano

    \item \textbf{Household Items}:\\
    computer\_keyboard, cup, dining\_table, eggnog, espresso, hot\_pot, lawn\_mower, plate, red\_wine, table\_lamp, teapot, television, throne, torch, tray, typewriter\_keyboard, wall\_clock, wardrobe, washer, water\_bottle, water\_jug, wine\_bottle, wok

    \item \textbf{Human Activities}:\\
    groom, scuba\_diver, tennis\_ball, volleyball
\end{itemize}

\begin{figure*}[h]
\begin{center}
\includegraphics[width=1\textwidth]{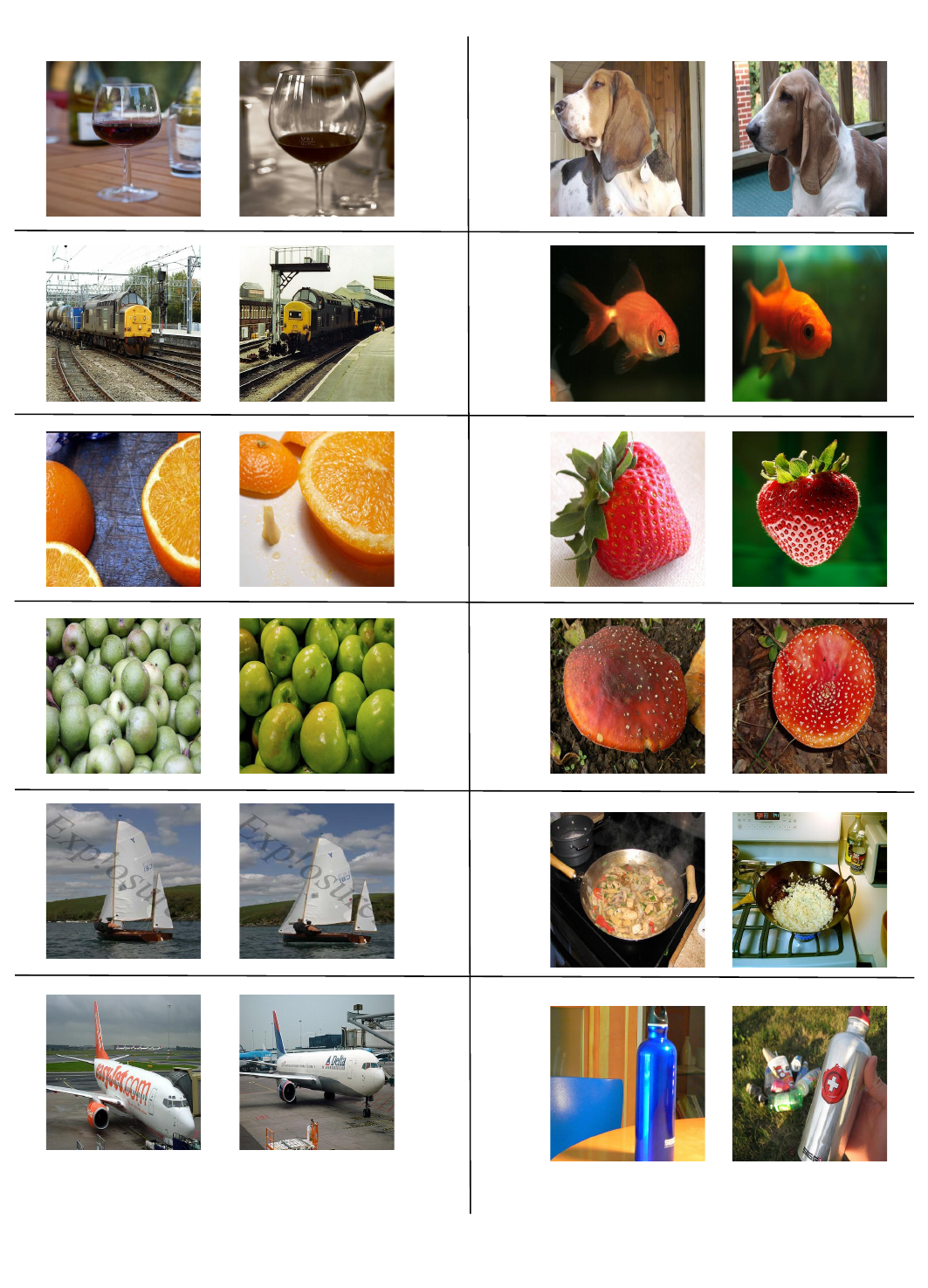}

\end{center}
   \vspace{-2 mm} 
   \caption{Examples from the semantic pairs dataset}
\label{fig:figure6}
\end{figure*}

\begin{figure*}[h]
\begin{center}
\includegraphics[width=1\textwidth]{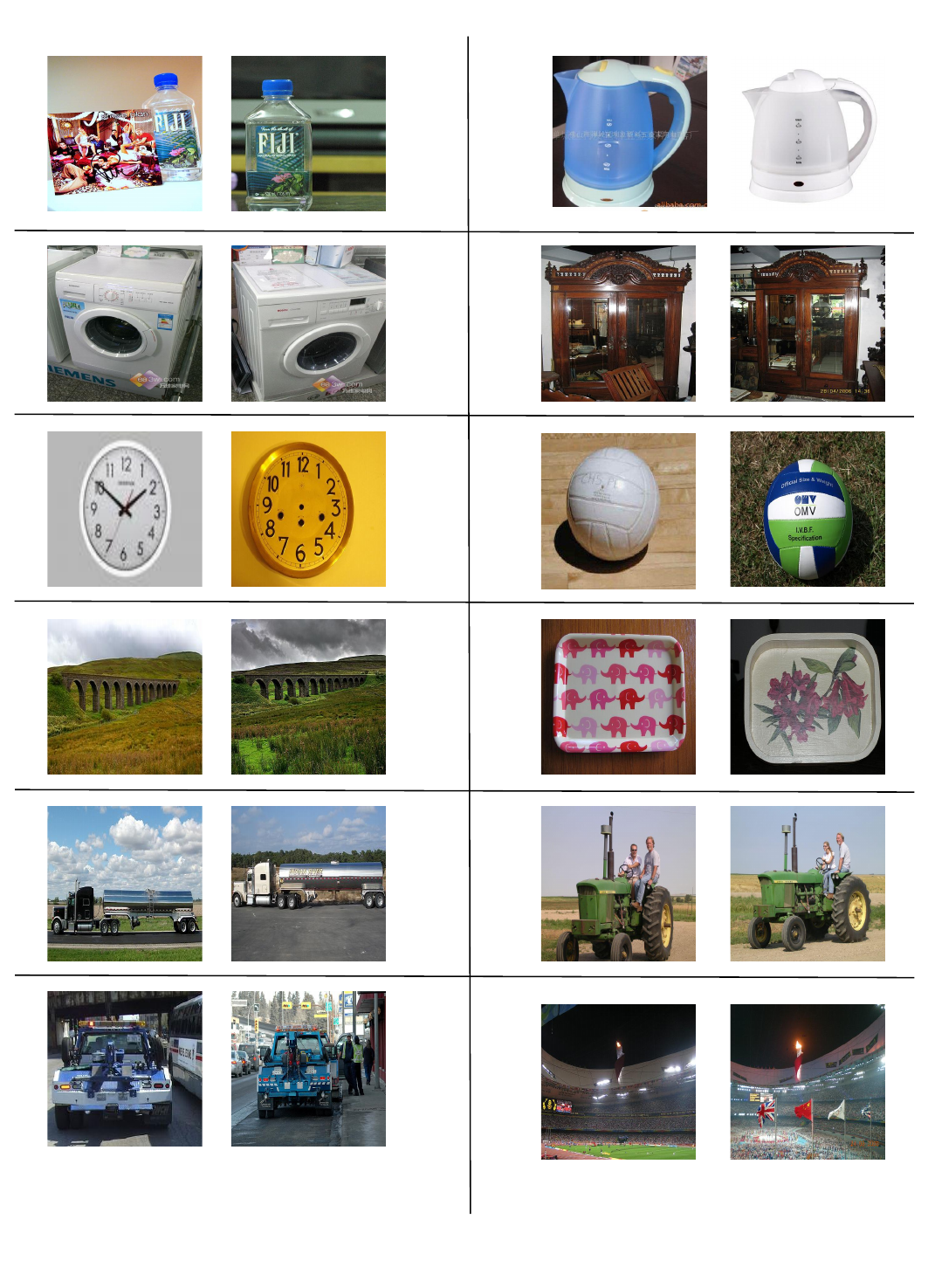}

\end{center}
 \vspace{-2 mm} 
   \caption{Additional examples from the semantic pairs dataset}
\label{fig:figure7}
\end{figure*}

\end{document}